\documentclass{article}
\usepackage[utf8]{inputenc}
\usepackage[margin = 1in]{geometry}
\usepackage{authblk}

\usepackage{amsmath, amssymb, amsthm, amsfonts, mathtools, multirow, mathrsfs, tabu, booktabs, lipsum, url, subcaption, graphicx,xcolor, tabu,  authblk, hyperref, blindtext}

\usepackage[normalem]{ulem}
\useunder{\uline}{\ul}{}

\usepackage[english]{babel}
\usepackage[shortlabels]{enumitem}
 \usepackage[ruled,vlined, linesnumbered]{algorithm2e}
\usepackage[capbesideposition=outside,capbesidesep=quad]{floatrow}
\usepackage[framemethod=tikz]{mdframed}
\usepackage{bbm}

\allowdisplaybreaks

\newcommand{\appropto}{\mathrel{\vcenter{
  \offinterlineskip\halign{\hfil$##$\cr
    \propto\cr\noalign{\kern2pt}\sim\cr\noalign{\kern-2pt}}}}}
    
\DeclareMathOperator*{\argmin}{argmin}

\usepackage[backend=biber,style=ieee,sorting=none,natbib=true]{biblatex} 
\usepackage[utf8]{inputenc}
\usepackage[autostyle=true, threshold=2]{csquotes}

\title{\textbf{Unsupervised Spatial-spectral Hyperspectral Image Reconstruction and Clustering with Diffusion Geometry}}
\author[1,2]{Kangning Cui}
\author[1]{Ruoning Li}
\author[3]{Sam L. Polk}
\author[3]{James M. Murphy}
\author[4]{Robert J. Plemmons}
\author[1,2]{Raymond H. Chan\footnote{Corresponding Author: raymond.chan@cityu.edu.hk}}
\affil[1]{ Department of Mathematics, City University of Hong Kong}
\affil[2]{ Hong Kong Centre for Cerebro-Cardiovascular Health Engineering}
\affil[3]{ Department of Mathematics, Tufts University}
\affil[4]{ Departments of Mathematics and Computer Science, Wake Forest University}
\date{}                     
\begin{document}
\topmargin=0mm

\maketitle
\begin{abstract}
Hyperspectral images, which store a hundred or more spectral bands of reflectance, have become an important data source in natural and social sciences.  
Hyperspectral images are often generated in large quantities at a relatively coarse spatial resolution. As such, unsupervised machine learning algorithms incorporating known structure in hyperspectral imagery are needed to analyze these images automatically.  
This work introduces the Spatial-Spectral Image Reconstruction and Clustering with Diffusion Geometry (DSIRC) algorithm for partitioning highly mixed hyperspectral images. 
DSIRC reduces measurement noise through a shape-adaptive reconstruction procedure. 
In particular, for each pixel, DSIRC locates spectrally correlated pixels within a data-adaptive spatial neighborhood and reconstructs that pixel's spectral signature using those of its neighbors.   
DSIRC then locates high-density, high-purity pixels far in diffusion distance (a data-dependent distance metric) from other high-density, high-purity pixels and treats these as cluster exemplars, giving each a unique label. Non-modal pixels are assigned the label of their diffusion distance-nearest neighbor of higher density and purity that is already labeled. 
Strong numerical results indicate that incorporating spatial information through image reconstruction substantially improves the performance of pixel-wise clustering. 
\end{abstract}


\section{Introduction}\label{sec:intro}

Hyperspectral images (HSIs) are high-dimensional images---often remotely sensed by airborne or orbital spectrometers---that encode rich spectral and spatial structure~\cite{bioucas2013hyperspectral} that has enabled the detection of material structure in a scene using machine learning algorithms~\cite{ADVIS, DVIS}. 
However, due to the large volume of HSI data continuously generated by remote sensors, expert annotations (often required for supervised algorithms) are usually difficult to obtain.  
Moreover, there is an inherent trade-off between spatial and spectral resolution in HSIs~\cite{bioucas2013hyperspectral, DVIS}.  HSIs are often created at a coarse spatial resolution due to this trade-off, meaning that some pixels in an HSI correspond to spatial regions in the scene containing many different materials~\cite{bioucas2013hyperspectral, DVIS}. Thus, it is crucial to develop unsupervised approaches that capture the underlying geometric structure of an HSI while incorporating spectral mixing. 

This work introduces the Spatial-Spectral Image Reconstruction and Clustering with Diffusion Geometry (DSIRC) algorithm for unsupervised material discrimination using HSIs. 
DSIRC is a variant of the unsupervised Diffusion and Volume maximization-based Image Clustering (D-VIC) algorithm~\cite{DVIS}.  DSIRC improves D-VIC by incorporating spatial information in Shape-adaptive Reconstruction (SaR), which smooths the spectral signatures of pixels and thus promotes local spatial regularity of HSIs before cluster analysis~\cite{LiSar2022}. 
Since HSI pixels that are spatially close tend to come from the same cluster, DSIRC substantially outperforms D-VIC (which is agnostic to spatial information) in extensive numerical results on real-world HSI data. 

This article is organized as follows. 
Section \ref{sec: background} contains background on HSI analysis (e.g., clustering, reconstruction, and spectral unmixing), diffusion geometry, and D-VIC. Section \ref{sec: DSIRC} introduces DSIRC. Section \ref{sec: numerics} contains numerical comparisons of DSIRC against classical and state-of-the-art algorithms. Section \ref{sec: conclusions} concludes and discusses future work.

\section{Related Works}
\label{sec: background}

\subsection{Hyperspectral Image Clustering}
Algorithms for HSI clustering segment HSI pixels  $X=\{x_i\}_{i=1}^N \subset~\mathbb{R}^B$ (interpreted as a point cloud of $N$ pixels' spectral signatures, where $B$ is the number of spectral bands) into \emph{clusters} of pixels $\{X_k\}_{k=1}^K$~\cite{friedman2001elements}. Ideally, any two pixels from the same cluster should share key commonalities (e.g., common materials~\cite{DVIS}). HSI clustering algorithms are \emph{unsupervised}; i.e., the partition $\{X_k\}_{k=1}^K$ is recovered without the aid of ground truth (GT) labels~\cite{DVIS}.

\subsection{Hyperspectral Image Reconstruction}

Spatially close HSI pixels tend to come from the same cluster, but intra-cluster spectral reflectances may vary substantially due to the inherently coarse spatial resolution of HSIs. This section reviews \emph{HSI reconstruction}, which efficiently denoises hyperspectral data by reconstructing HSI pixels using the spectra of spatial nearest neighbors. HSI reconstruction has been successfully used as a preprocessing step for semi-supervised learning~\cite{LiSar2022, LiMDPI} and is expected to be useful for unsupervised learning~\cite{DVIS}. 

HSI reconstruction algorithms denoise an image by reconstructing the spectral signature of each pixel $x\in X$ using a linear combination of spatial neighbors' spectral signatures.  
A pixel is considered a spatial neighbor of $x$ if it is contained in a spatial window centered at $x$ in the original image. 
While simple spatial squares have been successfully used as spatial windows in unsupervised and semi-supervised algorithms, the spatial radius generally requires tuning in practice~\cite{LiSar2022, LiMDPI, murphy2018unsupervised, murphy2019spectral, sam2021multi}.
In contrast, shape-adaptive (SA) regions may be used for parameter-free HSI reconstruction~\cite{LiSar2022}.

\subsection{Shape-adaptive Reconstruction}\label{sec: SaR}

This section introduces the SaR algorithm for HSI reconstruction. Denote the spatial coordinate of an HSI pixel $x\in X$ by $\xi_x=(\xi_1,\xi_2)^T$. We model the first principal component (PC) score of $x$~\cite{friedman2001elements}, denoted $\mathbf{Z}(\xi_x)$, as  $\mathbf{Z}(\xi_x) = {\mathbf{I}(\xi_x)} + \varepsilon_x$, where $\mathbf{I}(\xi_x)$ and ${\varepsilon_x}$ model the ideal signal and noise associated with the spectral signature $x$. 

To learn the SA region for $x$, SaR first estimates the signal $\mathbf{I}(\xi_x)$ using local polynomial approximation (LPA) filtering.  
Mathematically, for each direction $\theta_m$ ($m=~1,2,\dots,8$) and length candidate $l \in L_{sar}$ ($L_{sar}$ is the set of all possible length candidates), LPA estimates the ideal signal associated with $x$ using $\hat{\mathbf{I}}_{l,\theta_m}(\xi_x)=\sum_s{g_{l,\theta_m}(u_s)\mathbf{Z}(\mathbf{R}_{\theta_m}\xi_s)},$ where $g_{l,\theta_m}(u_s)$ is the directional LPA kernel~\cite{LiSar2022} for direction $\theta_m$ and length $l$,  $u_s=~\mathbf{R}_{\theta_m}(\xi_x-~\xi_s)$ is the rotated coordinate difference between $\xi_x$ and $\xi_s$ (each $\xi_s$ depends on $l$~\cite{katkovnik_2006}), 
and $\mathbf{R}_{\theta_m}=~{\begin{bmatrix}
\cos(\theta_m) & \sin(\theta_m) \\
-\sin(\theta_m) & \cos(\theta_m) \\
\end{bmatrix}}$. 
As such, LPA estimates the signal associated with the pixel $x$ for each direction $\theta_m$ and length candidate $l$. 

To select the optimal length $l_{\theta_m}^*\in L_{sar}$ for each direction $\theta_m$, SaR relies on the intersection of confidence intervals (ICI) rule, implemented on the first PC of $X$. Mathematically, for each direction $\theta_m$ and length candidate $l$, ICI estimates a confidence interval for the signal associated with the pixel $x$:  
$\text{CI}(\hat{\mathbf{I}}_{l,\theta_m}(\xi_x))=[\alpha_{l,\theta_m}(\xi_x),\beta_{l,\theta_m}(\xi_x)]$, using bounds $\alpha_{l,\theta_m}(\xi_x) = \hat{\mathbf{I}}_{l,\theta_m}-\tau \sigma\sum_s{g_{l,\theta_m}(u_s)^2}$ and $\beta_{l,\theta_m}(\xi_x)=\hat{\mathbf{I}}_{l,\theta_m}+\tau \sigma\sum_s{g_{l,\theta_m}(u_s)^2}$, and $\tau$ is a threshold that may be tuned via cross validation~\cite{katkovnik_2006}.
The length $l_{\theta_m}^*$ is then selected as the largest $l$ such that  $\max\limits_{i=1,\dots,j}{\alpha_{l_i,\theta_m}}\leq\min\limits_{i=1,\dots,j}{\beta_{l_i,\theta_m}}$~\cite{foi2007pointwise}.
In particular, for each direction $\theta_m$, ICI selects the largest length $l_{\theta_m}^*$ such that the intersection of confidence intervals of LPA-estimated signal values $\bigcap_{l_i\leq l}{\rm CI}(\hat{\mathbf{I}}_{l_i,\theta_m}(\xi_x))$ is nonempty~\cite{foi2007pointwise, fu2015hyperspectral}.

SaR uses the SA region learned through the procedure outlined above to reconstruct the  spectral signature of $x$ as $\tilde{x}=~\frac{\sum_{y\in Z(x)}w(x,y)y}{\sum_{y\in Z(x)}w(x,y)}$, where $Z(x)$ is the set of pixels contained in the SA region associated with $x$ and $w(x,y)$ is the Pearson correlation coefficient between $x$ and $y$~\cite{LiSar2022}. SaR (provided in Algorithm \ref{alg: SaR}) has been successfully used to aid in semi-supervised segmentation of HSIs~\cite{LiSar2022} and is expected to prove useful for unsupervised HSI clustering. 

\begin{algorithm}[t]
\SetAlgoLined
\KwIn{ $X$ (dataset), $L_{sar}$ (length candidates)}
\KwOut{$\tilde{X}$ (reconstructed dataset)}
Project pixel spectra on to their first PC~\cite{friedman2001elements}\;
\For{$x\in X$}{
Estimate the signal associated with $x$ using LPA for each length $l\in L_{sar}$ and direction $\theta_m$ ($m=1,2,\dots 8$)~\cite{LiSar2022}\;
For each direction $\theta_m$ ($m=1,2,\dots, 8$), compute the optimal length $l_{\theta_m}^*$ for that direction using ICI~\cite{foi2007pointwise}. Denote the pixels in the resulting SA region as $Z(x)$\; 
Reconstruct $x$ as
$\tilde{x} = \frac{\sum_{y\in Z(x)}w(x,y)y}{\sum_{y\in Z(x)}w(x,y)}$\;
} 
\caption{Shape-adaptive Reconstruction (SaR)}\label{alg: SaR}
\end{algorithm}

\subsection{Blind Spectral Unmixing}

Due to an inherent tension between spatial and spectral resolution, HSIs are often generated at a coarse spatial resolution~\cite{bioucas2013hyperspectral}. As such, a single pixel may correspond to a spatial region containing multiple materials~\cite{DVIS,bioucas2008HySime}.
Linear spectral unmixing algorithms recover latent material structure in HSIs by decomposing pixel spectra into a linear combination of endmembers: spectral signatures intrinsic to the materials in the scene. 
Mathematically, if $p$ is the number of materials in the scene, a blind spectral unmixing algorithm locates a matrix $\mathbf{E}\in\mathbb{R}^{p\times B}$ (with rows containing endmembers) and abundance vectors $a_i\in\mathbb{R}^p$ such that  $x_i \approx a_i^{\top}\mathbf{E}$~\cite{bioucas2013hyperspectral,cui2021unsupervised}.
The \emph{purity} of a pixel $x_i$---defined by  $\eta(x_i)=\max_{1\leq j\leq p}{ } (a_i)_j$---therefore quantifies the level of mixture in the pixel $x_i$. Indeed, $\eta(x)$ will be large only if it corresponds to a spatial region containing predominantly just one material~\cite{ADVIS, DVIS,cui2021unsupervised}.

\subsection{Diffusion Geometry}

Graph-based clustering methods efficiently extract latent nonlinear structure in HSIs by interpreting pixels as nodes in an undirected, weighted graph~\cite{coifman2006diffusion}. Edges between nodes are encoded in an adjacency matrix $\mathbf{W}\in\mathbb{R}^{N\times N}$; $\mathbf{W}_{ij}=1$ if $x_j$ is one of the $k_{n}$ $\ell^2$-nearest neighbors of $x_i$ in $X$, and  $\mathbf{W}_{ij}=0$ otherwise. Let $\mathbf{D}$ be the $N\times N$ diagonal degree matrix with $\mathbf{D}_{ii} = \sum_{j=1}^N \mathbf{W}_{ij}$. Then, $\mathbf{P} = \mathbf{D}^{-1}\mathbf{W}$ may be interpreted as the transition matrix for a Markov diffusion process on HSI pixels. If the graph underlying $\mathbf{P}$ is irreducible and aperiodic, then $\mathbf{P}$ has a unique stationary distribution $\pi\in\mathbb{R}^{1\times N}$ satisfying $\pi \mathbf{P}=\pi$.

\emph{Diffusion distances} enable comparisons between pixels in the context of the diffusion process encoded in $\mathbf{P}$. Define the diffusion distance at time $t\geq 0$ between $x_i,x_j\in X$ by $D_t(x_i, x_j) = \sqrt{\sum_{k=1}^N [(\mathbf{P}^t)_{ik}-(\mathbf{P}^t)_{jk}]^2/\pi_k }$~\cite{coifman2006diffusion}. The diffusion time parameter $t$ controls the scale of structure considered by diffusion distances; smaller $t$ corresponds to the recovery of local structure and larger $t$ corresponds to the recovery of global structure~\cite{sam2021multi,murphy2021multiscale}. Diffusion distances may be efficiently computed using the eigendecomposition of $\mathbf{P}$. Indeed, if $\{(\lambda_k, \psi_k)\}_{k=1}^N$ are the (right) eigenvalue-eigenvector pairs of the transition matrix $\mathbf{P}$, then $D_t(x_i,x_j) = \sqrt{\sum_{k=1}^N|\lambda_k|^{2t}[(\psi_k)_i -(\psi_k)_j]^2}$ for any $t\geq 0$ and $x_i,x_j\in X$.
Importantly, for $t$ sufficiently large, diffusion distances therefore can be accurately approximated by using just the eigenvectors $\psi_k$ with sufficiently large $|\lambda_k|$.

\subsection{Diffusion and Volume Maximization-based Image Clustering} \label{sec: D-VIC}

D-VIC is a highly-accurate diffusion-based HSI clustering algorithm that directly incorporates spectral mixing into its labeling procedure~\cite{DVIS}. D-VIC first estimates $\eta(x)$ through a standard spectral unmixing step: using  Hyperspectral Subspace Identification by Minimum Error to estimate $p$, Alternating Volume Maximization to estimate $\mathbf{E}$, and a nonnegative least square solver to estimate abundances~\cite{bioucas2008HySime,chan2011simplex, bro1997fast}.  
Empirical density of pixels is estimated using $f(x) = 
\sum_{y\in NN_{k_n}(x)}\exp(-\|x-y\|_2^2/\sigma_0^2)$, where $NN_{k_n}(x)$ is the set of $k_n$ $\ell^2$-nearest neighbors of $x$ in $X$ and $\sigma_0>0$ is the scaling factor controlling the interaction radius between pixels in density calculations. 
Denoting $\zeta(x)$ as the harmonic mean of normalized purity $\hat{\eta}(x)=\frac{\eta(x)}{\max_{1\leq i\leq N}\eta(x_i)}$ and density $\hat{f}(x)=\frac{f(x)}{\max_{1\leq i\leq N}f(x_i)}$, the following function is constructed to incorporate diffusion geometry into mode selection:
\begin{align*}
    d_t(x) = 
    \begin{cases} 
        \max\limits_{y\in X}D_t(x,y) & x = \argmin\limits_{y\in X}\zeta(y),\\
        \min\limits_{y\in X}\{D_t(x,y)| \zeta(y)\geq \zeta(x)\} & \text{otherwise.} 
    \end{cases}
\end{align*}
By definition, the $K$ maximizers of $\mathcal{D}_t(x)=\zeta(x)d_t(x)$ are pixels high in density and purity but far in diffusion distance from other pixels high in density and purity. These pixels are selected as class modes and given unique labels. 
Non-modal pixels are (in order of non-increasing $\zeta(x)$) assigned the label of their labeled $D_t$-nearest neighbor of higher $\zeta$-value that is already labeled.

\section{Spatial-Spectral Image Reconstruction and Clustering with Diffusion Geometry}
\label{sec: DSIRC}

Real-world HSIs often encode strong spatial regularity; i.e., pixels that are spatially close are more likely to contain the same materials. 
As such, incorporating the rich spatial structure contained in HSIs often leads to substantially higher performance among algorithms for HSI clustering and segmentation~\cite{LiSar2022,LiMDPI,murphy2018unsupervised,murphy2019spectral,sam2021multi}.

\begin{algorithm}[tb]
\SetAlgoLined
\KwIn{ $X$ (dataset), $k_n$ (\# nearest neighbors), \\  $\sigma_0$ (KDE bandwidth),   $t$ (diffusion time parameter),\\ $K$ (\# classes), $L_{sar}$ (length candidates)}
\KwOut{$\hat{\mathcal{C}}$ (HSI clustering)}
For each $x\in X$, compute $\eta(x)$ and $f(x)$\; 
For each $x\in X$, compute $\zeta(x) = \frac{2\hat{f}(x) \hat{\eta}(x)}{\hat{f}(x)+ \hat{\eta}(x)}$\;
Compute the reconstructed data $\tilde{X}={\rm SaR}(X)$\;
Label $\hat{\mathcal{C}}(\tilde{x}_{m_k}) = k$ for $1\leq k \leq K$, where $\{\tilde{x}_{m_k}\}_{k=1}^K$ are the $K$ maximizers of  $\mathcal{D}_t(\tilde{x}) = \zeta(x) d_t(\tilde{x})$\;
In order of non-increasing $\zeta(x)$, assign each non-modal $\tilde{x}\in \tilde{X}$ the label $\hat{\mathcal{C}}(\tilde{x}) = \hat{\mathcal{C}}(\tilde{x}^*)$,  where $\tilde{x}^* = \argmin\limits_{y\in X}\{D_t(\tilde{x},y)|\zeta(y)\geq \zeta(\tilde{x}) \ \land \ \hat{\mathcal{C}}(y)>0\}$\;
\caption{Spatial-Spectral Image Reconstruction and Clustering with Diffusion Geometry}\label{alg: DSIRC}
\end{algorithm} 

This section introduces the DSIRC clustering algorithm, which explicitly incorporates HSI reconstruction into D-VIC. In its first stage, DSIRC computes $\zeta(x)$ using the original pixel spectra (as is described in Section \ref{sec: D-VIC})~\cite{DVIS}.
Then, the SaR algorithm (Section \ref{sec: SaR}) is implemented on $X$: a denoising step that incorporates spatial information into its HSI reconstruction~\cite{LiSar2022}.  Diffusion distances are calculated using the SaR-reconstructed image. The $K$ maximizers of $\mathcal{D}_t(\tilde{x})=\zeta(x)d_t(\tilde{x})$ are considered cluster modes and assigned unique labels. Non-modal pixels are (in order of non-increasing $\zeta(x)$) assigned the label of their labeled $D_t$-nearest neighbor of higher $\zeta$-value that is already labeled. As such, the sole difference between DSIRC and D-VIC is that DSIRC incorporates spatial information through its SaR-based HSI reconstruction step, and D-VIC is agnostic to spatial information~\cite{DVIS,LiSar2022}.

\begin{figure*}[ht]
    \centering
    
    \begin{subfigure}[t]{0.19\textwidth}
    \centering
    \includegraphics[width = \textwidth]{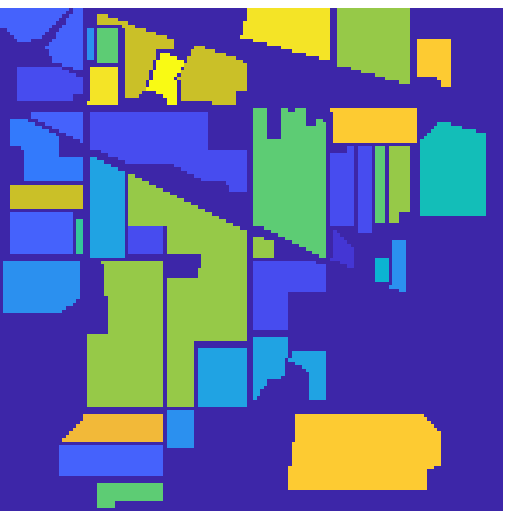} \hspace{0.02in}
    \vspace{-0.5cm}
    \caption{Ground Truth}
    \end{subfigure}
    \begin{subfigure}[t]{0.19\textwidth}
    \centering
    \includegraphics[width = \textwidth]{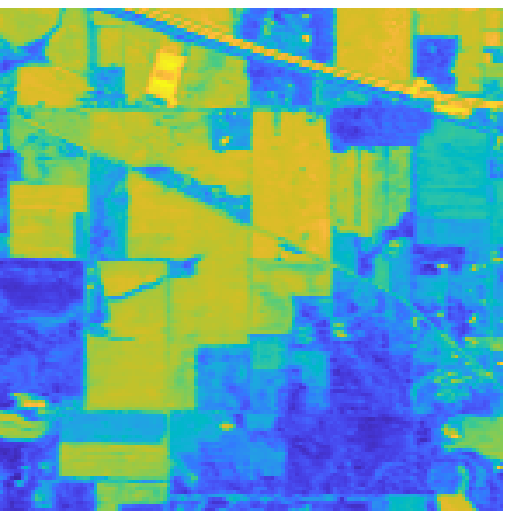} \hspace{0.02in}
    \vspace{-0.5cm}
    \caption{First PC}
    \end{subfigure}
    \begin{subfigure}[t]{0.19\textwidth}
    \centering
    \includegraphics[width = \textwidth]{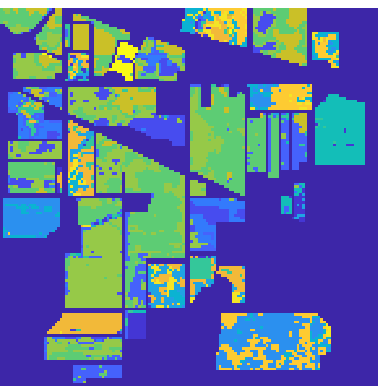} \hspace{0.02in}
    \vspace{-0.5cm}
    \caption{$K$-Means}
    \end{subfigure}
    \begin{subfigure}[t]{0.19\textwidth}
    \centering
    \includegraphics[width = \textwidth]{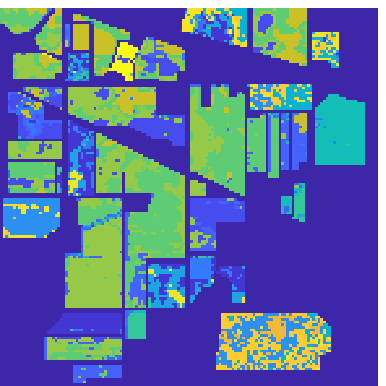} \hspace{0.02in}
    \vspace{-0.5cm}
    \caption{$K$-Means PCA}
    \end{subfigure}
    \begin{subfigure}[t]{0.19\textwidth}
    \centering
    \includegraphics[width = \textwidth]{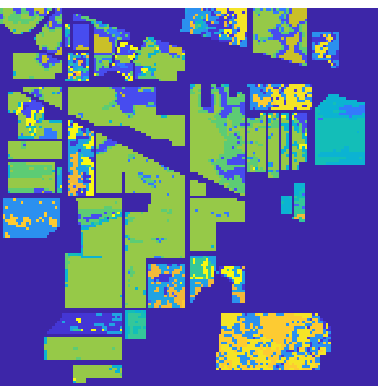} 
    \vspace{-0.5cm}
    \caption{GMM PCA}
    \end{subfigure}
    
    \begin{subfigure}[t]{0.19\textwidth}
    \centering
    \includegraphics[width = \textwidth]{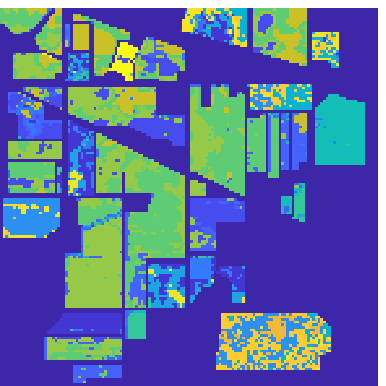} \hspace{0.02in}
    \vspace{-0.5cm}
    \caption{SC}
    \end{subfigure}
    \begin{subfigure}[t]{0.19\textwidth}
    \centering
    \includegraphics[width = \textwidth]{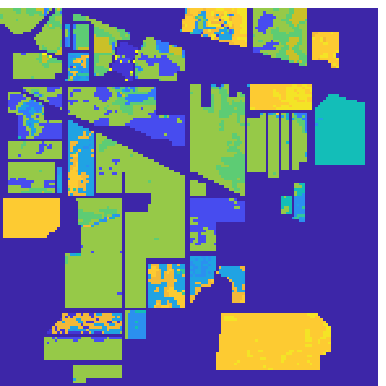} \hspace{0.02in}
    \vspace{-0.5cm}
    \caption{D-VIC}
    \end{subfigure}
    \begin{subfigure}[t]{0.19\textwidth}
    \centering
    \includegraphics[width = \textwidth]{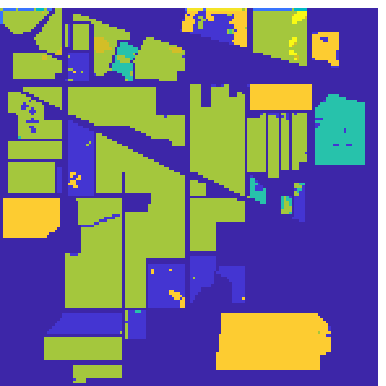} \hspace{0.02in}
    \vspace{-0.5cm}
    \caption{SC-I}
    \end{subfigure}
    \begin{subfigure}[t]{0.19\textwidth}
    \centering
    \includegraphics[width = \textwidth]{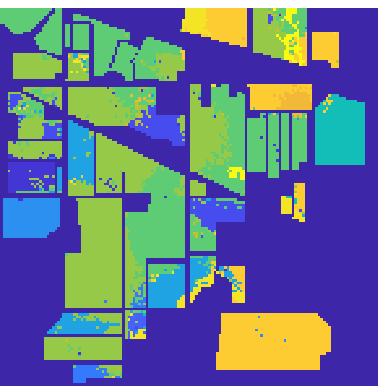} \hspace{0.02in}
    \vspace{-0.5cm}
    \caption{DLSS}
    \end{subfigure}
    \begin{subfigure}[t]{0.19\textwidth}
    \centering
    \includegraphics[width = \textwidth]{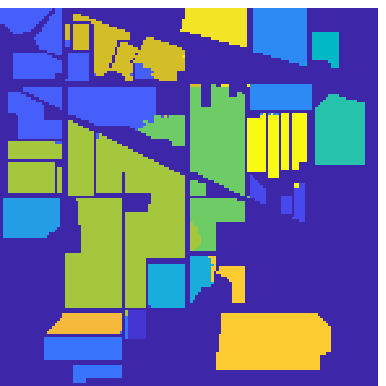} 
    \vspace{-0.5cm}
    \caption{DSIRC}
    \end{subfigure}
    
    \caption{Comparison of clusterings results of classical algorithms (Panels (c)-(e)), state-of-the-art algorithms (Panels (f)-(i)) and DSIRC (Panel (j)) on the Indian Pines HSI (Panel (a)-(b)). DSIRC---the only algorithm evaluated that incorporates spatial information into a diffusion-based clustering scheme that accounts for spectral mixing---produces substantially better HSI segmentations than comparison methods evaluated. }
    \label{fig:results}
\end{figure*}

\section{Experimental Results}\label{sec: numerics}

This section contains comparisons of DSIRC against related HSI clustering algorithms on the Indian Pines HSI.  
Indian Pines---collected by the NASA AVIRIS sensor in  northwest Indiana, USA---encodes $B=200$ bands of reflectance  across $145\times 145$ pixels. 
The Indian Pines scene consists of $K=16$ GT classes, which are visualized in Fig.\ref{fig:results}(a). 
Fig.\ref{fig:results}(b) visualizes the first PC of Indian Pines. 
Clusterings were evaluated using overall accuracy (OA)---the fraction of correctly labeled pixels---and Cohen's $\kappa$-coefficient $\kappa=\frac{{\rm OA}-p_e}{1-p_e}$, ($p_e$ is the probability of random agreement between two labelings). 

The classical algorithms we compared DSIRC against were $K$-Means and the Gaussian Mixture Model (GMM)~\cite{friedman2001elements}.
$K$-Means locates the clustering that minimizes within-cluster Euclidean distances to cluster centroids. GMM fits a mixture of Gaussian distributions to the dataset using the expectation-maximization algorithm. 
As HSIs have hundreds of spectral bands, PCA dimensionality reduction is often implemented before cluster analysis using $K$-Means and GMM~\cite{friedman2001elements}.

We also compared against several state-of-the-art graph-based clustering algorithms. To emphasize the improvement associated with incorporating spatial information, we evaluated two graph-based algorithms that are agnostic to spatial information: spectral clustering (SC)~\cite{ng2001spectral} and D-VIC (see Section \ref{sec: D-VIC})~\cite{DVIS}.
SC implements $K$-Means after the nonlinear mapping $x_i\mapsto[(\psi_1)_i\;(\psi_2)_i\dots (\psi_K)_i]$~\cite{ng2001spectral}. We also compared DSIRC against graph-based clustering algorithms that use spatial information. First considered was improved spectral clustering (SC-I), which modifies the graph underlying $\mathbf{P}$ in SC to incorporate spatial information into edge weights~\cite{zhao2019fast}. Additionally, we evaluated spectral-spatial diffusion learning (DLSS), which incorporates spatial information into a graph-based clustering framework~\cite{maggioni2019learning} by restricting edges between pixels to spatial nearest neighbors in a $(2R+1)\times (2R+1)$ spatial square centered at those pixels, where $R$ is a tunable spatial window input parameter~\cite{murphy2018unsupervised}. 
We optimized for OA across the same hyperparameter grid for all graph-based algorithms.  
The set of length candidates for DLSS and DSIRC ranged across the same set: $L_{sar} = \{1,2,3,5,7,9\}$. 

Table \ref{tab:results} compares the performance of DSIRC against the methods described above, and Figure \ref{fig:results} visualizes Indian Pines and optimal clusterings obtained by DSIRC and comparison methods.  
DSIRC outperformed DLSS (its closest competitor) by 0.13 in OA, and 0.21 in $\kappa$.
The main difference between these two algorithms is that DSRIC incorporates pixel purity into its mode selection and utilizes the spatial regularity of the HSI before its unsupervised diffusion-based labeling process. 
In contrast, DLSS incorporates spatial information through a spatially-regularized graph but does not directly rely on pixel purity to label the HSI~\cite{murphy2018unsupervised}.
Furthermore, DSIRC relies on a spatially-adaptive window with automatically-determined shape, whereas DLSS requires the user to input the spatial window size $R$~\cite{murphy2018unsupervised}.
Image reconstruction in DSIRC appears to efficiently remove ``spatial noise'' observed in the D-VIC clustering, as is visualized in Figure \ref{fig:results}. 
Thus, enforcing spatial regularity appears to improve the quality of a diffusion-based clustering quite substantially.

\begin{table}[tb]
\centering
\resizebox{0.6\textwidth}{!}{%
\begin{tabular}{|c|cc|c|cc|}
\hline
                       & \textbf{OA} & \boldmath{$\kappa$} &    & \textbf{OA} & \boldmath{$\kappa$} \\ \hline
\textbf{GMM PCA}                  & 0.3581      & 0.2821     & \textbf{SC-I}  & 0.4696          & 0.3493          \\
\textbf{SC}                       & 0.3784      & 0.3029     & \textbf{D-VIC} & 0.4756          & 0.3848          \\
\boldmath{$K$}\textbf{-Means}     & 0.3817      & 0.3080     & \textbf{DLSS}  & {\ul 0.4886}          & {\ul 0.4074}          \\
\boldmath{$K$}\textbf{-Means PCA} & 0.3837      & 0.3085     & \textbf{DSIRC} & \textbf{0.6195} & \textbf{0.6123} \\ \hline
\end{tabular}%
}
\caption{Performances of DSIRC and comparison methods on the Indian Pines dataset. Bold and underlined values indicate highest and second-highest performances, respectively. Performances of $K$-Means, $K$-Means PCA, GMM PCA, SC, D-VIC, and DSIRC were averaged across ten trials. DSIRC achieved substantially higher performance than any comparison method.}
\label{tab:results}
\end{table}

\section{Conclusions and Future Works}\label{sec: conclusions}
We conclude that incorporating spatial information through image reconstruction appears to substantially improve the performance of pixel-wise HSI clustering algorithms that exploit known HSI structure such as spectral mixing. Thus, incorporating a shape-adaptive reconstruction akin to that which was used in DSIRC may be useful before the labeling of HSI pixels. 
Future work includes extending DSIRC to the active learning domain, wherein the labels of a few carefully-selected pixels are queried and propagated to the rest of the image~\cite{ADVIS,murphy2018unsupervised}. We also expect that DSIRC may be extended to the unsupervised multiscale clustering setting~\cite{sam2021multi,murphy2021multiscale}. The resulting unsupervised and active learning algorithms are likely to be successful in a number of applications, e.g., identifying changes of mining ponds in multispectral images over time, possibly reflecting the occurrence of artisanal and small-scale gold mining activities~\cite{SedaChange2022}.


\printbibliography 

\end{document}